# A New 4-DOF Robot for Rehabilitation of Knee and Ankle-Foot Complex: Simulation and Experiment

Afshin Alipour[a,1,†], Mohammad J. Mahjoob[a,b,†,*], Zahra Fakhari[c], and Ara Nazarian[b,d]

[a] *School of Mechanical Engineering, University of Tehran, Tehran, Iran*

[b] *BIDMC, Harvard Medical School, Boston, USA*

[c] *Rehabilitation Department, Tehran University of Medical Sciences, Tehran, Iran*

[d] *Department of Orthopaedic Surgery, Yerevan State Medical University, Yerevan, Armenia*

[1] *New affiliation: Erik Jonsson School of Engineering and Computer Science, University of Texas at Dallas, Richardson, USA*

[†] *These authors have contributed equally to this work.*

[*] *Corresponding author: mmahjoob@ut.ac.ir*

**Abstract**

Stationary robotic trainers are lower limb rehab robots which often incorporate an exoskeleton attached to a stationary base. The issue observed in the stationery trainers for simultaneous knee and ankle-foot complex joints is that they restrict the natural motion of ankle-foot in the rehab trainings due to the insufficient Degrees of Freedom (DOFs) of these trainers. A new stationary knee-ankle-foot rehab robot with all necessary DOFs is developed here. A typical rehab training is first implemented in simulation, and then tested on a healthy subject. Results show that the proposed system functions naturally and meets the requirements of the desired rehab training.

## 1. Introduction

Damage to the nervous system, caused by accidents such as stroke and spinal cord injuries, often leads to movement disorders [1]. This issue is of practical importance, as it can severely hamper activities of daily living for the survivors, in light of the large number of stroke incidents per year [2]. To address this issue, patients take rehabilitation exercises under the supervision of therapists to regain their abilities.

During the last two decades, rehabilitation exercises have received significant attention from robotic researchers. Many robots have been designed to facilitate rehab trainings for both patients and therapists, and to improve the training results in terms of repeatability, reliability and accuracy in evaluating the patient's progress [3]. Among the developed platforms, lower limb rehab robots are of more importance, since they have a direct effect on gait.

The lower limb rehab robots can be divided into two general types: standing/walking and sitting/lying. The standing/walking robots mainly aim to correct the gait pattern in conditions similar to daily life. This type of robot brings three groups under its umbrella. The first group includes treadmill gait trainers that incorporate a treadmill and an exoskeleton attached to the patient's leg to adjust the motion of the joints, while a bodyweight support usually holds a portion of the patient's weight [4], [5]. The second group consists of foot-plate-based gait trainers, which use two footplates attached to the foot sole [6], [7]. These robots simulate walking in different conditions such as walking on an even surface or climbing up/down the stairs. The third group contains the over-ground gait trainers which help patients to do rehabilitation in daily life or after surgery [8], [9]. Active orthoses are also categorized as over-ground robots. Their primary purpose is to correct the lower limb orientation during gait. They are mostly passive and must only brace the injured joint. However, active orthoses have also been developed to further improve the patients' gait [10], [11].

Sitting/lying rehab robots (also called stationary-based trainers) are often comprised of an exoskeleton attached to a stationary frame. Patients sit or lie next to the device and strap their legs to the exoskeleton. They are mainly used at early stages of the rehab process when the patient needs to increase Range of Motion (ROM) of joints and strengthen her/his muscles [12]. Hence, rehab training applied with this type of robots is conducted in different modes. One of the generally used modes is passive assistance (also known as Continuous Passive Motion (CPM)) during which an external force is applied to move the joints across the existing range, while the patient exerts no force [13], [14]. Although passive motion helps to decrease the joints stiffness and increase ROM [15], it has been observed that the active participation of patients in rehab trainings can improve the results [16].

Many stationary-based robots rehabilitate the hip, knee or ankle separately. For instance, hip and knee joints are mainly addressed by robots with one active Degree of Freedom (DOF) in the sagittal plane to allow Flexion/Extension (F/E) motion [17]–[19]. On the other hand, rehab robots focusing on ankle joint usually provide more than one DOF to account for the complex kinematics of the ankle-foot complex [20]–[24]. Some of them use a fixed central strut which puts a center of rotation on the foot sole for ankle joint, which is completely unrealistic and inconsistent with the nature of the ankle-foot complex [20], [23]. However, simultaneous rehabilitation of joints is of interest. Patients are required by therapists to move joints simultaneously during specific types of stretch trainings [25]. Besides, joints work together during daily activities. In fact, the training which involves several joints at the same time appears to provide a more physiologically relevant condition. Therefore, several robotic platforms were introduced to incorporate more than one human joint [26]–[32].

The problem with the existing mechanisms which simultaneously rehabilitate more than one joint is their lack of necessary DOFs for natural motion of the joints, especially for the ankle-foot complex where only the ankle Plantarflexion/Dorsiflexion (P/D) is addressed [26], [28], [30]–[33]. These robots usually perform rehab training in the sagittal plane. Nonetheless, it has been shown that even the ankle P/D requires the other two anatomical planes as well [34]–[36]. Given the other important DOF of the ankle-foot complex as Inversion/Eversion (I/E), the existing designs do not provide the ankle-foot complex with its own natural pattern. To the best of our knowledge, all robots dealing with the knee and the ankle consist of only knee F/E along with ankle P/D.

A new mechanism capable of producing all necessary motions is thus of prime interest. The robot presented here generates such motions based on the natural DOFs of the knee and the ankle-foot complex. This portable robot employs electrical actuators for the robot to be practically used at home as well. Also, its DOFs are precisely measured by encoders aligned with the rotational axes of human joints. This approach avoids the use of inverse kinematics existing (inevitably) in parallel-mechanism rehab robots [20]–[23] that causes complexity and errors when simplifications are made in the relations.

The paper is organized as follows. The necessary DOFs for knee and ankle-foot rehabilitation are explained first. The configuration of the robot is presented in the next section. Simulations are then carried out to estimate the required motor torques for the desired rehab training and to synthesize the appropriate control loop. Results and Discussion section then follow the Simulations section.

## 2. Necessary DOFs for the rehabilitation

There are two DOFs for the knee joint [37]. the first one is F/E in which the shank rotates around the knee in the sagittal plane. The second DOF is Internal/External Rotation (IR/ER) which can be realized only when the knee is flexed. In this DOF, the shank rotates in the transverse plane around its central axis passing through the knee and the ankle. Since IR/ER's contribution to activities of daily living is relatively small, F/E is the only DOF chosen for rehab trainings. For the ankle and foot motions to be easily realized, Kapandji studied them in three different anatomical planes shown in Figure 1 [37]. The rotation around x-axis is called P/D which is mainly produced by the ankle joint [34]. This motion is generated by the rotation of the foot around the ankle in the sagittal plane. However, the axis of this rotation is slightly oblique in practice. This motion is strongly recommended in rehab trainings.

Foot rotation around the z-axis (along the shank), produces Abduction-Adduction (A/A) in the transverse plane. Ankle and subtalar joints contribute almost equally to this motion [34]. Foot also rotates around the y-axis, i.e. along the foot, to produce Supination/Pronation (S/P) in the coronal plane, which is mainly generated at the subtalar joint. Kapandji also expresses that A/A and S/P coexist with each other due to the foot structure [37]. It means if motion is applied to one DOF (i.e. A/A), the foot structure causes the other DOF to move as well (i.e. S/P). As a matter of fact, a combination of adduction, supination and slight plantarflexion makes a unique motion called inversion. On the other hand, a combination of abduction, pronation, and slight dorsiflexion generates eversion. I/E is also important in rehab protocols [38]. Therefore, the ankle-shank complex has two DOFs: ankle P/D and ankle-foot I/E. Since I/E already consists of P/D, one can conclude that knee F/E and ankle-foot I/E are the important DOFs considered in rehab trainings. Therefore, our robot must be able to generate these motions as described.

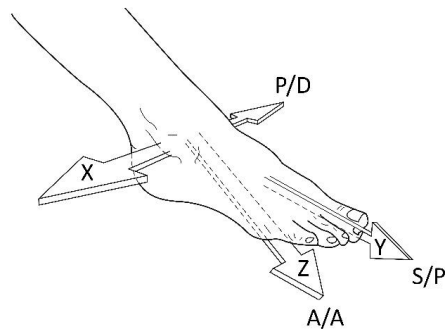

Figure 1. Ankle-foot complex's rotational axes (adopted from Ref. [37]).

## 3. The proposed robot

According to the previous section, the robot should be able to move the shank around the knee F/E axis perpendicular to the sagittal plane. For ankle-foot I/E, the robot may have a single rotation around the I/E axis. However, this is quite impractical. Firstly, the orientation of the I/E axis changes during the course of I/E motion [39]. Secondly, the I/E axis can be highly subjective, since it may vary from person to person. An alternative approach is that the robot may separately generate P/D, A/A and S/P rotations perpendicular to the sagittal, transverse and coronal planes, respectively. Then, I/E can be achieved by controlling the motions in the aforementioned planes. Besides, the robot can accommodate everyone by adjusting motion parameters.

Therefore, the robot must be able to rotate the foot in the sagittal plane around the shank to generate P/D, in the coronal plane along the foot length for S/P, and in the transverse plane for A/A. Despite F/E, P/D and A/A, the location of the S/P axis is rather controversial. Some researchers have reported that this axis passes through the sole [40], while others believe that it passes through the ankle [24].

To address this controversy, one can determine the location of S/P based on the kinematics of the ankle-foot complex. Given this kinematics, an optimization algorithm can then be applied to accurately determine the location of the robot's S/P axis to help the foot hold its own natural motion during I/E. One common method for obtaining this kinematics is to use motion capture systems [41].

However, due to the lack of data for ankle-foot complex in the literature, we take the benefit of the two-hinge-joint model presented in the literature [35]. In this model, the ankle-foot complex is modeled with three rigid bodies: shank, talus, and the foot. Then, P/D and I/E motions are given by two fixed hinge joints connecting the shank to the talus (talocrural joint), and the talus to the foot (subtalar joint), as shown in Figure 2a. We use this model developed in the literature to generate the data points for the optimization step.

The position of the foot as a rigid body can be obtained from three points on the sole: $P$, $M$, and $N$ as in Figure 2a. Then, the foot is moved based on the two-hinge-joint model to generate P/D and I/E at the same time. Considering the ankle point (point $B$ in Figure 2a) as a fixed origin, this motion can be described using trajectories of the aforementioned three points, as illustrated in

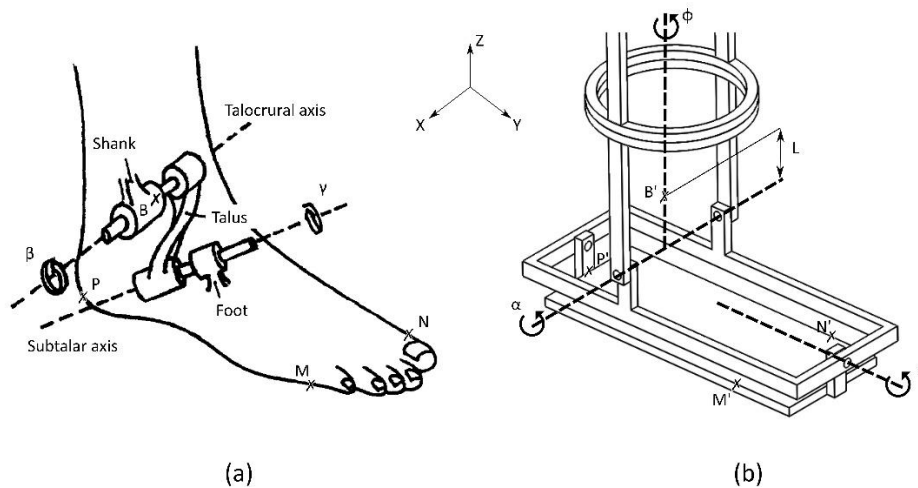

Figure 2. a) The two-hinge-joint model for ankle-foot complex (adopted from Ref. [50]) b) robot model.

Figure 3 for an average human with 1.7 m height. We pick eleven points from each trajectory in Figure 3 to use as the target points in our optimization algorithm.

Then, the location of the S/P axis can be calculated so that the robot can closely generate the P/D and I/E motions. A schematic of the robot is shown in Figure 2b which will be discussed with details later in this section. The robot provides P/D, S/P and A/A motions by rotations around $X$, $Y$ and $Z$ axes with values of $\alpha$, $\psi$ and $\varphi$, respectively. These rotations are realized with revolute joints for P/D and S/P, and with two sliding rings placed around the shank for A/A. Here, the S/P axis is shown with an arbitrary distance of $L$ below the point $B'$ which corresponds to the ankle point $B$ when the patient leg is placed in the robot. The goal here is to make $P'$, $M'$, and $N'$ points on the robot closely track $P$, $M$, and $N$ trajectories during P/D and I/E (Figure 3). For this aim, eleven points are selected from the trace of each of $P'$, $M'$, and $N'$ points, and are called design points: $P'_i$, $M'_i$ and $N'_i$ where $i = 1, ..., 11$. As an example, the coordinates of $P'_i$ can be calculated directly from the following equation:

$$P'_i = \begin{bmatrix} \cos\varphi_i & -\sin\varphi_i & 0 \\ \sin\varphi_i & \cos\varphi_i & 0 \\ 0 & 0 & 1 \end{bmatrix} \begin{bmatrix} 1 & 0 & 0 \\ 0 & \cos\alpha_i & -\sin\alpha_i \\ 0 & \sin\alpha_i & \cos\alpha_i \end{bmatrix} \left( \begin{bmatrix} \cos\psi_i & 0 & \sin\psi_i \\ 0 & 1 & 0 \\ -\sin\psi_i & 0 & \cos\psi_i \end{bmatrix} \left( P'_0 - \begin{bmatrix} 0 \\ 0 \\ -L \end{bmatrix} \right) + \begin{bmatrix} 0 \\ 0 \\ -L \end{bmatrix} \right)$$
(1)

where $P'_0$ is the initial coordinates of $P'$ when all the robot rotations are zero. Then, the vector of design variables, $W$, will consist of 34 elements as:

$$W = [L, \varphi_1, \varphi_2, ..., \varphi_{11}, \alpha_1, \alpha_2, ..., \alpha_{11}, \psi_1, \psi_2, ..., \psi_{11}]$$
(2)

Here, the cost function is considered to be the average Euclidean distance between the eleven target points and corresponding design points. Therefore, the optimization problem can be formulated as:

$$\min_W f_{cost} = \min_W \frac{1}{11} \sum_{i=1}^{11} (\|P_i - P'_i\|_2 + \|M_i - M'_i\|_2 + \|N_i - N'_i\|_2)$$
(3)

*Subject to*:

$w_i \in [lb_i, ub_i]$

where $lb_i$ and $ub_i$ are the lower and upper bounds of design variables.

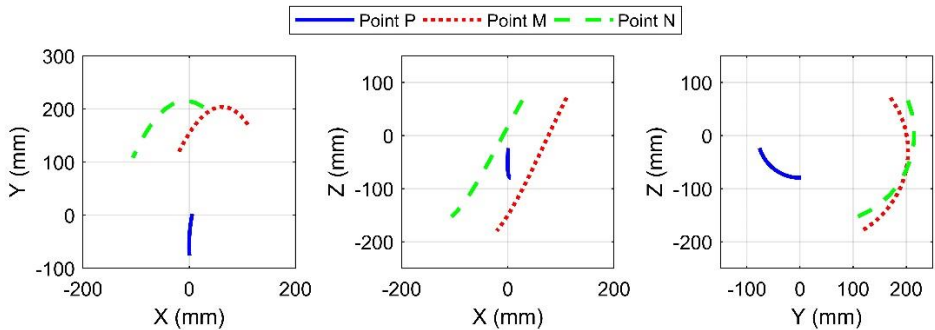

Figure 3. Trajectories of target points during I/E motion (considering the ankle point as a fixed origin.)

Due to the non-linearity of the problem, an evolutionary algorithm (Differential Evolution, DE), is chosen to solve the problem [42]. The speed of convergence and ease of application have made DE a great candidate for designing mechanisms [43]. DE starts with an initial population. Each individual in this population, $W_{i,G}$ where $i = 1, \dots, NP$, is randomly chosen within the design space and is a potential solution to the problem. $NP$ and $G$ are population size and generation number, respectively, which need to be tuned for each problem. For this work, $NP$ is set to 100.

After determining the best individual in the population with the smallest cost value as $W_{best,G}$, a perturbation vector is obtained for each individual as follows:

$$V_i = W_{best,G} + F(W_{r_1} - W_{r_2}) \tag{4}$$

where $r_1$ and $r_2$ are two randomly chosen indices from $1, \dots, NP$ different form $i$ and $best$. $F$ is the amplification of differential variation, set to 0.6 for this work.

Before moving to the next step, mutation is also applied here with a probability of 0.1 in order to prevent stagnation in local minimums [43]. Then, $V_i$ and $W_{i,G}$ are crossed with probability of 0.8 to produce a trial vector, $u_i$. If $u_i$ has smaller cost function than $W_{i,G}$, it will replace $W_{i,G}$ in the next generation, otherwise $W_{i,G}$ will remain unchanged. This algorithm will continue until the maximum number of generations is reached.

After running the optimization algorithm for a sufficient number of generations, the optimum value of variable $L$ converges to 0.2 mm below the ankle point with an optimum cost value of 0.8 mm. Considering the simplifications, approximations, and errors made in the problem, the outcome of the optimization step suggests that the S/P axis of the robot should pass through the ankle point. In conclusion, the proposed robot should provide the rotation of foot around the ankle point in coronal plane.

To validate the results of the optimization step, we conducted a practical test before the fabrication of the final prototype. Therefore, we fabricated three simple prototypes to validate the optimization result (Figure 4). In these prototypes, S/P goes through the ankle, the middle of the ankle and sole, and the sole. A subject wore the prototypes and moved his foot in the I/E pattern. As expected from the optimization step, the first prototype in which the S/P axis goes through the ankle turned out to be compatible best with the I/E pattern, because only in this prototype, the rings forming A/A remained in contact with each other. Hence, the final prototype of the exoskeleton was fabricated based on this mechanism.

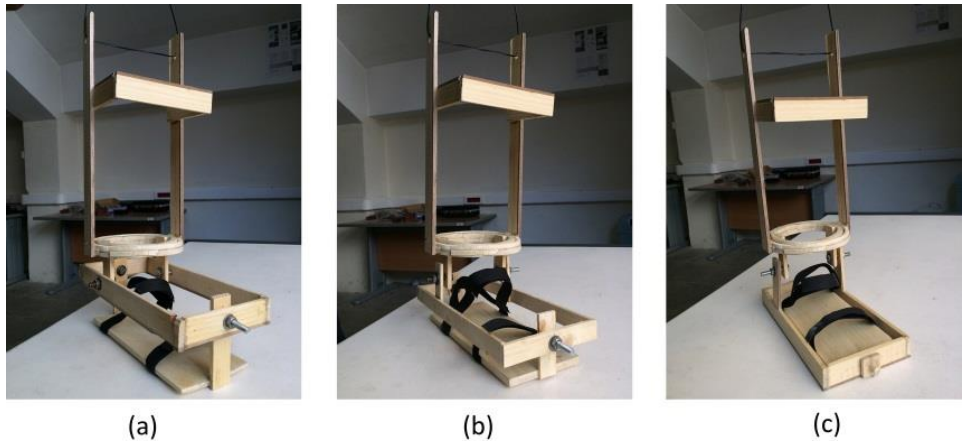

Figure 4. Different mechanisms for the exoskeleton: S/P passing through (a) ankle (b) the middle of the ankle and sole (c) the sole.

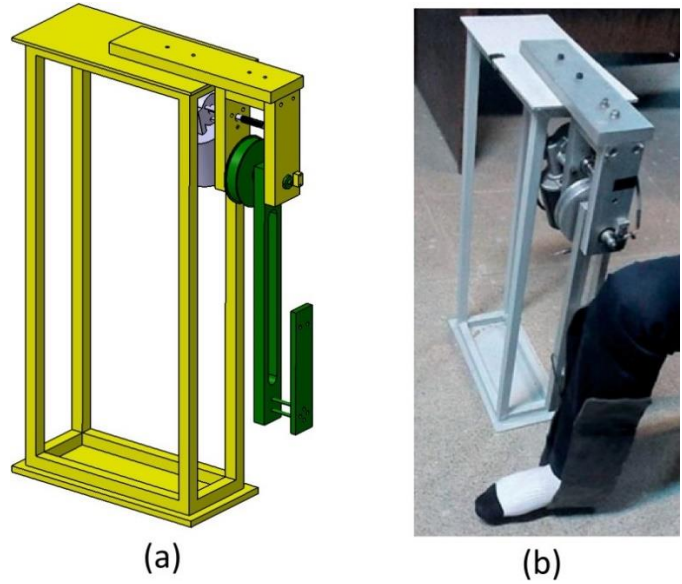

Figure 5. Knee rehab robot: (a) Computer-Aided Design (CAD) model (b) final prototype [19].

In the meantime, a stationary-based robot for knee rehabilitation was previously designed and fabricated in our laboratory [19]. This robot has one DOF for knee F/E (Figure 5) and consists of a metal stationary base, an aluminium movable link to which the shank is strapped, and a DC motor ($\theta$-motor). A cable-driven transmission system with a ratio of 1:10 is incorporated to transmit power from $\theta$-motor to the link, and a 200-pulse optical encoder attached to the link's rotational shaft measures the knee F/E. The patient sits/lies next to the robot with her/his shank strapped to the link. Then, a control system comprised of a microcontroller, a motor driver and a current sensor helps to move the shank in a desired trajectory in the sagittal plane.

In order to realize a 4-DOF knee and ankle-foot-complex rehabilitation robot, the link of the previous robot is replaced with the fabricated final prototype of the exoskeleton (Figure 6). The exoskeleton consists of four main parts: upper shank (*us*), lower shank (*ls*), foot frame (*ff*) and foot plate (*fp*). The '*us*' directly replaces the previous link and has only one rotation ($\theta$) with respect to the base in the sagittal plane with a revolute joint for knee F/E. The '*ls*' also has one rotation ($\phi$) with respect to the '*us*' in the transverse plane with a custom-made two-way thrust ball bearing, which can also take radial loads to produce A/A. The '*ff*' rotates around the ankle in the sagittal plane also with a revolute joint to generate ankle P/D ($\alpha$). Finally, foot S/P ($\psi$) is the only motion of the '*fp*' with respect to the '*ff*' produced by another revolute joint in the coronal plane.

Generally, the robot must fit patients with different statures. Therefore, the '*fp*' is considered big enough to fit large feet. Besides, the lengths of '*fp*' and '*us*' are made adjustable. By these arrangements, patients with statures ranging from 1.6 m to 1.9 m can use the final prototype. The robot is also compatible with both legs. If one wishes to use the robot for her/his left leg, the '*ls*' should turn 180° around the shank.

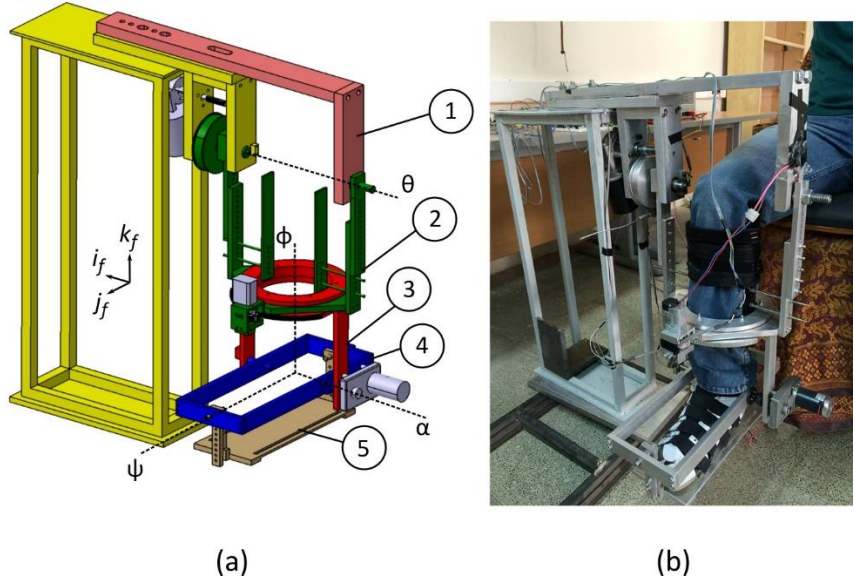

Figure 6. The proposed knee and ankle-foot rehab robot: (a) CAD model: 1- stationary base 2- '*us*' 3- '*ls*' 4- '*ff*' 5- '*fp*' (b) The leg position in the robot.

Furthermore, the ROMs of the robot's DOFs are satisfactory. In an average human, the knee can be flexed up to 140° [37]. Ankle P/D can change between 30° in dorsiflexion to 40° in plantarflexion. A/A varies from 30° of adduction to 20° of abduction, and the supination and pronation are limited to 50° and 30°, respectively. In this robot, θ easily satisfies the knee F/E requirement, ϕ is restricted to ±60°, α can vary from 60° in dorsiflexion to 70° in plantarflexion, and ψ is constricted to ±50° due to the robot structure.

Also, some rehab protocols need the robot to be active. During CPM, for example, patients exert no force, therefore external actuation is required. The knee F/E is thus activated with a DC motor (θ-motor) equipped with a cable-driven power transmission system (θ-power transmission system) in the previous work. Ankle P/D also has to be actuated. A DC motor (α-motor) fixed to the '*ls*' is used to move the '*ff*' directly. However, according to [37], A/A and S/P do not need to be actuated simultaneously through their interdependent existence. During preliminary experiments, we decided that A/A should be actuated because S/P was found more dependent on A/A than the reverse. For this reason, a motor (ϕ-motor) was fixed to the '*us*', and its power transmitted to the '*ls*' by a pinion-gear power transmission system (ϕ-power transmission system) with the ratio of 1:7.5. The pinion was attached to the motor shaft, while the gear was made by milling gear teeth around the custom-made thrust ball bearing.

Figure 7 shows the components of the robot. Four 200-pulse rotary optical encoders (HEDS-9700-E50, Avago Technologies) are employed to measure rotations. The θ-encoder detecting knee F/E is placed on the '*us*' rotational shaft. The data from ϕ-encoder put on the ϕ-motor shaft are divided by 7.5 to get the real value of A/A. The α- and ψ- encoders are directly located on the P/D and S/P rotational shafts. The patient's leg is strapped to the exoskeleton from two points (Figure 6 b). First, the shank is strapped to the '*us*' to ensure that it has the same motion as the '*us*'. Second, a shoe attached to the '*fp*' holds the foot and makes sure that rotations recorded by encoders are the real rotations of the foot.

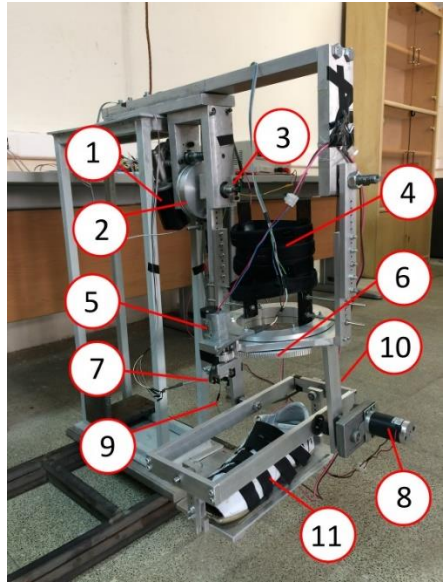

Figure 7. Components of the robot: 1- θ-motor 2- θ-power transmission system 3- θ-encoder 4- the shank straps 5- ϕ-motor 6- ϕ-power transmission system 7- ϕ-encoder 8- α-motor 9- α-encoder 10- ψ-encoder (behind the link) 11- the shoe.

## 4. Simulations

The first aim here is to estimate the required torques to drive the robot which also helps us to select appropriate motors. Furthermore, the rehab training considered to be implemented in experiments, i.e. passive assistance, is simulated in MATLAB to obtain the control loop requirements before practical tests. A dynamic model representing the system is therefore required.

The system is composed of the robot and a human. Obtaining the dynamic model of human lower limbs is challenging and has been studied with different points of view. The prevalent method is to represent anatomical elements such as muscles and tendons with mechanical entities, for instance mass, stiffness, and viscosity. In this case, the slope of the applied external force versus elongation is used to define the stiffness of these biological elements. However, Latash and Zatsiorsky [44] showed that stiffness assessed in this way conveys not only the mechanical properties of the biological element but also the experiment procedure itself. That is why the reported values for stiffness vary from work to work. They also pointed out that viscoelastic properties can apply to the biological limbs only when they are passive.

Besides, considering all biological elements such as muscles, bones, tendons, ligaments, joint capsules, etc. in the dynamic model will dramatically increase the complexity of the model. To tackle this problem, some researchers have tried to use simplistic mechanisms inspired by the physics of the actual lower limbs [45]. To obtain a simplistic model in the present study, we considered the foot, shank, and thigh as rigid bodies connecting to each other with mechanical joints. In fact, the ankle and the knee are assumed to resemble spherical and revolute joints, respectively. Also, we model the effect of other biological tissues, e.g. muscles, tendons, skin, ligaments, etc., on the passive motion of the lower limbs as rotational springs and dampers acting on the joints.

The values of stiffness and viscosity being used in our model must be already assessed in a similar experiment condition. Since we intend to implement the passive assistance training, the viscoelastic properties of the knee joint assessed with the pendulum test seem suitable. In the pendulum test, the patient sits on a chair while her/his shank is hung freely. Then, the clinician brings the patient's shank up while the thigh remains still and then releases it. By measuring the free oscillations of the knee, the viscoelastic properties can be calculated. So, we considered a linear rotational spring and damper with values of 3.58 Nm.rad$^{-1}$ and 0.1 Nms.rad$^{-1}$ respectively, which were obtained with a pendulum test [46].

Similarly, the passive stiffness of the ankle and the ankle-foot complex are conventionally measured by passively moving the foot in the anatomical planes, which is similar to the procedure in passive assistance training. Then, by calculating the slope of the external torque versus the angular motion, the stiffness can be calculated. Chen at al [47] assessed the passive stiffnesses of the ankle and the ankle-foot complex in the three anatomical planes which suits the spherical joint for the ankle in our model. We approximated the reported stiffnesses in all planes with fourth-order polynomials in Nm.Radian$^{-1}$ as in equations (5), (6) and (7) for ankle P/D, ankle-foot A/A and S/P, respectively. Figure 8 shows the ankle and the ankle-foot complex stiffnesses in all planes along with their approximated fourth-order polynomials. As it can be seen, the joints show a rather compliance behaviour towards plantarflexion, adduction and supination directions.

$$K_\alpha = 313.9\alpha^4 + 203.2\alpha^3 + 59.3\alpha^2 + 13.8\alpha + 6.6 \tag{5}$$

$$K_\varphi = 3163.5\varphi^4 - 554.7\varphi^3 - 112.5\varphi^2 + 7.3\varphi + 7.6 \tag{6}$$

$$K_\psi = 5010.1\psi^4 - 1739.5\psi^3 + 402.1\psi^2 - 17.1\psi + 11.2 \tag{7}$$

Here, the coordinates are considered as follows: when the knee is flexed and the shank is perpendicular to the thigh, θ is assumed to be zero. If the foot is in its natural position, and it is perpendicular to the shank, ϕ, α and ψ are also considered zero. The displacements of θ, ϕ, α and ψ are assumed positive for knee extension, foot adduction, ankle dorsiflexion and foot supination, respectively. Five coordinate frames are used to describe relative motions: $\mathcal{F}, \mathcal{A}, \mathcal{B}, \mathcal{C}$ and $\mathcal{D}$. The frame $\mathcal{F}$ is the reference frame which is attached to the stationary base and has no motion. While $\mathcal{A}$ is fixed to the '*us*' and has a simple angular velocity ($\dot{\theta}$) with respect to $\mathcal{F}$. Similarly, $\mathcal{B}, \mathcal{C}$ and $\mathcal{D}$ are affixed to the '*ls*', '*ff*' and '*fp*' and have simple angular velocities ($\dot{\varphi}, \dot{\alpha}$ and $\dot{\psi}$) with respect to $\mathcal{A}$, $\mathcal{B}$ and $\mathcal{C}$, respectively. To express vectors in each of these frames, dextral sets of orthogonal unit vectors of $(\hat{\imath}_f,\hat{\jmath}_f,\hat{k}_f)$, $(\hat{\imath}_a,\hat{\jmath}_a,\hat{k}_a)$, $(\hat{\imath}_b,\hat{\jmath}_b,\hat{k}_b)$, $(\hat{\imath}_c,\hat{\jmath}_c,\hat{k}_c)$ and $(\hat{\imath}_d,\hat{\jmath}_d,\hat{k}_d)$ are chosen in frames $\mathcal{F}$, $\mathcal{A}$, $\mathcal{B}$, $\mathcal{C}$ and $\mathcal{D}$, respectively. These sets of unit vectors are in the same direction when all generalized coordinates are zero ($\theta=\phi=\alpha=\psi=0$). Figure 6a shows the first set.

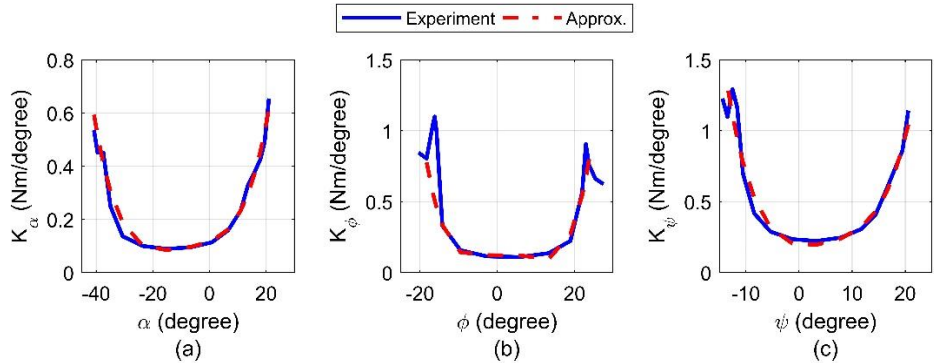

Figure 8. The ankle and ankle-foot complex stiffness obtained from experiments [47] and its approximation in the three anatomical planes: a) P/D stiffness b) A/A stiffness c) S/P stiffness.

The next step is to determine the system parameters. Mechanical properties of the human foot and shank, e.g. masses, position of center of masses (COMs) and moments of inertia for an average person with 1.7 m height and 65 Kg mass are calculated by anthropometric tables [48]. The data which were not reported are estimated. The mechanical properties of the robot are obtained from the CAD model. However, the patient shank ('*s*') has the exact motion of the '*us*' in practice. Therefore, the physical properties of the '*s*' can be added to the '*us*' and reported as a single body ('*us-s*'). This phenomenon also exists between the patient foot ('*f*') and the '*fp*'. Therefore, they can also be treated as a single body: '*fp-f*'. The mechanical properties of the system are now reported in Table 1.

### 4.1 Desired torques

Here, the motion equations are derived by Lagrange's method. The required torques for a sample trajectory are then obtained and based on them, appropriate actuators are selected. To derive the equations, the angular velocities of the links in $\mathcal{F}$ are calculated as follows:

$$\vec{\omega}^{\mathcal{A}/\mathcal{F}} = \dot{\theta}\hat{\imath}_a$$
$$\vec{\omega}^{\mathcal{B}/\mathcal{F}} = \vec{\omega}^{\mathcal{B}/\mathcal{A}} + \vec{\omega}^{\mathcal{A}/\mathcal{F}} = \dot{\phi}\hat{k}_b + \dot{\theta}\hat{\imath}_a$$
$$\vec{\omega}^{\mathcal{C}/\mathcal{F}} = \vec{\omega}^{\mathcal{C}/\mathcal{B}} + \vec{\omega}^{\mathcal{B}/\mathcal{F}} = \dot{\alpha}\hat{\imath}_c + \dot{\phi}\hat{k}_b + \dot{\theta}\hat{\imath}_a$$
$$\vec{\omega}^{\mathcal{D}/\mathcal{F}} = \vec{\omega}^{\mathcal{D}/\mathcal{C}} + \vec{\omega}^{\mathcal{C}/\mathcal{F}} = \dot{\psi}\hat{\jmath}_d + \dot{\alpha}\hat{\imath}_c + \dot{\phi}\hat{k}_b + \dot{\theta}\hat{\imath}_a \qquad (8)$$

Table 1. Mechanical properties of the system. COMs are measured when all generalized coordinates are zero, considering knee as the origin.

| Property | Upper shank and shank (*us-s*) | Lower shank (*ls*) | Foot frame (*ff*) | Foot and footplate (*fp-f*) |
|---|---|---|---|---|
| Mass (kg) | 5.072 | 2.034 | 0.958 | 1.666 |
| $X_{COM}$ (cm) | 2.1 | -3.8 | -0.1 | 0 |
| $Y_{COM}$ (cm) | 0.8 | -1.1 | 4.5 | 6.7 |
| $Z_{COM}$ (cm) | -14.7 | -29.7 | -40 | -45.2 |
| $I_{xx}$ (kg.cm$^2$) | 780 | 170 | 120 | 170 |
| $I_{yy}$ (kg.cm$^2$) | 1050 | 280 | 40 | 20 |
| $I_{zz}$ (kg.cm$^2$) | 510 | 220 | 190 | 160 |
| $I_{xy}$ (kg.cm$^2$) | 0 | -20 | 0 | 0 |
| $I_{xz}$ (kg.cm$^2$) | -160 | -80 | 0 | 0 |
| $I_{yz}$ (kg.cm$^2$) | 30 | -20 | 0 | 0 |

$\vec{\omega}^{A/\mathcal{F}}$, $\vec{\omega}^{B/\mathcal{F}}$, $\vec{\omega}^{C/\mathcal{F}}$ and $\vec{\omega}^{D/\mathcal{F}}$ are the angular velocities of the '*us-s*', '*ls*', '*ff*' and '*fp-f*', respectively. The velocities of the COMs in $\mathcal{F}$ are calculated as:

$$\vec{V}^{C_{us-s}/\mathcal{F}} = \vec{\omega}^{A/\mathcal{F}} \times \vec{R}^{OC_{us-s}}$$

$$\vec{V}^{C_{ls}/\mathcal{F}} = \vec{\omega}^{B/\mathcal{F}} \times \vec{R}^{OC_{ls}}$$

$$\vec{V}^{C_{ff}/\mathcal{F}} = \vec{\omega}^{C/\mathcal{F}} \times \vec{R}^{OC_{ff}}$$

$$\vec{V}^{C_{fp-f}/\mathcal{F}} = \vec{\omega}^{D/\mathcal{F}} \times \vec{R}^{OC_{fp-f}} \tag{9}$$

$C_{us-s}$, $C_{ls}$, $C_{ff}$ and $C_{fp-f}$ are the COMs of the '*us-s*', '*ls*', '*ff*' and '*fp-f*', respectively. $\vec{V}^{C_{us-s}/\mathcal{F}}$, $\vec{V}^{C_{ls}/\mathcal{F}}$, $\vec{V}^{C_{ff}/\mathcal{F}}$ and $\vec{V}^{C_{fp-f}/\mathcal{F}}$ are the linear velocities of the COMs in $\mathcal{F}$. $\vec{R}^{OC_{us-s}}$, $\vec{R}^{OC_{ls}}$, $\vec{R}^{OC_{ff}}$ and $\vec{R}^{OC_{fp-f}}$ are position vectors from the knee representing point, $O$, to the COMs. The kinetic energy of the system ($KE$) is derived below:

$$KE = \frac{1}{2}m_{us-s}\vec{V}^{C_{us-s}/\mathcal{F}} \cdot \vec{V}^{C_{us-s}/\mathcal{F}} + \frac{1}{2}\vec{\omega}^{A/\mathcal{F}} \cdot I_{us-s}\vec{\omega}^{A/\mathcal{F}} + \frac{1}{2}m_{ls}\vec{V}^{C_{ls}/\mathcal{F}} \cdot \vec{V}^{C_{ls}/\mathcal{F}} + \frac{1}{2}\vec{\omega}^{B/\mathcal{F}} \cdot I_{ls}\vec{\omega}^{B/\mathcal{F}} + \frac{1}{2}m_{ff}\vec{V}^{C_{ff}/\mathcal{F}} \cdot \vec{V}^{C_{ff}/\mathcal{F}} + \frac{1}{2}\vec{\omega}^{C/\mathcal{F}} \cdot I_{ff}\vec{\omega}^{C/\mathcal{F}} + \frac{1}{2}m_{fp-f}\vec{V}^{C_{fp-f}/\mathcal{F}} \cdot \vec{V}^{C_{fp-f}/\mathcal{F}} + \frac{1}{2}\vec{\omega}^{D/\mathcal{F}} \cdot I_{fp-f}\vec{\omega}^{D/\mathcal{F}} \tag{10}$$

In the above equation, $m_{us-s}$, $m_{ls}$, $m_{ff}$ and $m_{fp-f}$ are the masses and $I_{us-s}$, $I_{ls}$, $I_{ff}$ and $I_{fp-f}$ are the inertia matrix of the links. If we refer to $g$ as the gravitational acceleration, and $z_{us-s}$, $z_{ls}$, $z_{ff}$ and $z_{fp-f}$ as the vertical distance between the point $O$ and the COMs in $\mathcal{F}$, then the potential energy of the system ($PE$) can be obtained by (11).

$$PE = m_{us-s}gz_{us-s} + m_{ls}gz_{ls} + m_{ff}gz_{ff} + m_{fp-f}gz_{fp-f} + \int_0^\theta K_\theta t\, dt + \int_0^\alpha K_\alpha t\, dt + \int_0^\varphi K_\varphi t\, dt + \int_0^\psi K_\psi t\, dt \tag{11}$$

The Lagrangian ($L$) is then calculated:

$$L = KE - PE \tag{12}$$

External torques ($\vec{T}_i$) applied to the system are as expressed below in equation (13). The rehab training candidate plan to test the robot is the passive assistance. Therefore, the subject does not exert any voluntary force. So, the torques in (13) are applied by motors or the viscous damping (except $T_\psi$).

$$\vec{T}_1 = T_\theta \hat{\imath}_a$$

$$\vec{T}_2 = T_d \hat{\imath}_a$$

$$\vec{T}_3 = T_\varphi \hat{k}_b$$

$$\vec{T}_4 = -T_\varphi \hat{k}_b$$

$$\vec{T}_5 = T_\alpha \hat{\imath}_c$$

$$\vec{T}_6 = -T_\alpha \hat{\imath}_c$$

$$\vec{T}_7 = T_\psi \hat{\jmath}_d$$

$$\vec{T}_8 = -T_\psi \hat{j}_d \tag{13}$$

$\vec{T}_1$ is the vector of the θ-motor's torque ($T_\theta$) applied on to the 'us-s'. $\vec{T}_2$ is the vector of the damping viscous ($T_d$) exerted on the 'ls'. $\vec{T}_3$ is the vector of the ϕ-motor's torque ($T_\varphi$) exerted on the 'ls', while $\vec{T}_4$ is its reaction on the 'us-s'. $\vec{T}_5$ is the vector of the α-motor's torque ($T_\alpha$) applied on the 'ff', and $\vec{T}_6$ is its reaction on the 'ls'. Since S/P is automatically produced by the foot to adjust the foot orientation when the A/A exists, a hypothetical torque, $T_\psi$, is assumed to apply on the 'fp-f' through the $\vec{T}_7$ vector. Its reaction is $\vec{T}_8$ exerted on the 'ff'. The angular velocities of the links ($\vec{\Omega}_i$), on which the external torques are applying, are presented in equation (14).

$$\begin{aligned}
\vec{\Omega}_1 &= \vec{\omega}^{A/F} \\
\vec{\Omega}_2 &= \vec{\omega}^{A/F} \\
\vec{\Omega}_3 &= \vec{\omega}^{B/F} \\
\vec{\Omega}_4 &= \vec{\omega}^{A/F} \\
\vec{\Omega}_5 &= \vec{\omega}^{C/F} \\
\vec{\Omega}_6 &= \vec{\omega}^{B/F} \\
\vec{\Omega}_7 &= \vec{\omega}^{D/F} \\
\vec{\Omega}_8 &= \vec{\omega}^{C/F}
\end{aligned} \tag{14}$$

Active external loads applied on each generalized coordinate ($Q_\theta$, $Q_\varphi$, $Q_\alpha$ and $Q_\psi$) are computed as follows:

$$\begin{aligned}
Q_\theta &= \sum_{i=1}^{8} \vec{T}_i \cdot \frac{d\vec{\Omega}_i}{d\dot{\theta}} \\
Q_\varphi &= \sum_{i=1}^{8} \vec{T}_i \cdot \frac{d\vec{\Omega}_i}{d\dot{\varphi}} \\
Q_\alpha &= \sum_{i=1}^{8} \vec{T}_i \cdot \frac{d\vec{\Omega}_i}{d\dot{\alpha}} \\
Q_\psi &= \sum_{i=1}^{8} \vec{T}_i \cdot \frac{d\vec{\Omega}_i}{d\dot{\psi}}
\end{aligned} \tag{15}$$

Equation (16) describes the motion derived by Lagrange's method. As long as the system has four generalized coordinates, we have four differential equations in (16) representing the dynamics of the system, while the actual system has three DOFs. Later in the 'Control Synthesis' subsection, the constraint between S/P and A/A will be used to reduce the number of equations to three.

$$\begin{aligned}
\frac{d}{dt}\left(\frac{dL}{d\dot{\theta}}\right) - \frac{dL}{d\theta} &= Q_\theta \\
\frac{d}{dt}\left(\frac{dL}{d\dot{\varphi}}\right) - \frac{dL}{d\varphi} &= Q_\varphi \\
\frac{d}{dt}\left(\frac{dL}{d\dot{\alpha}}\right) - \frac{dL}{d\alpha} &= Q_\alpha
\end{aligned}$$

$$\frac{d}{dt}\left(\frac{dL}{d\dot\psi}\right) - \frac{dL}{d\psi} = Q_\psi \tag{16}$$

If the generalized coordinates in (16) are known, the only unknown parameters in these equations are $T_\theta$, $T_\varphi$, $T_\alpha$ and $T_\psi$. Hence, to obtain an estimate for the required torques of the robot, equations in (16) are solved for the four torques. Then a reference trajectory is considered as in (17). In this sample trajectory, the right leg moves between two states (sinusoidal) with a period of 15 seconds. In the first state, the knee is flexed and ankle-foot is in inversion, and in the second state, the knee is extended and ankle-foot is in eversion.

$$\theta = 45(1 - \cos(2\pi t/15))$$
$$\varphi = 15\cos(2\pi t/15)$$
$$\alpha = -20\cos(2\pi t/15)$$
$$\psi = 10\cos(2\pi t/15) \tag{17}$$

To validate the above results (obtained via Lagrange's equations), the required torques for the sample trajectory are also obtained by using ADAMS software. The comparison between the torques from mathematical model and the ones from ADAMS (Figure 9) confirms that the equations obtained in (16) form an acceptable mathematical model of the system. Therefore, we can use its results to select appropriate motors, and then the verified dynamic model is used in the control loop simulations.

*4.2 Control Synthesis*

In this section, the desired rehab training is simulated to synthesize the control-loop before the experiments. Based on the progress of the patient during the rehab process, different trainings can be implemented by stationary-based robots [49]. These trainings are categorized into three main groups: assistive, strengthening, and proprioceptive. Assistive training is a very common practice to recover the lost ROM at the early stages of rehabilitation. It can be done in both passive and active modes. During passive assistance (or CPM), the robot must move the patient's leg in a specified trajectory cyclically to retrieve the ROMs of joints. The reference trajectory in this training is determined by therapists. Therefore, a position feedback control strategy is needed to ensure that the robot follows the reference path and does not go beyond the safe ROMs. On the other hand, in the active mode, the patients are not able to complete this task themselves, therefore the robot helps them by applying assistive forces. For this type of training, an impedance or admittance control scheme in conjunction with position and force feedback loops are required.

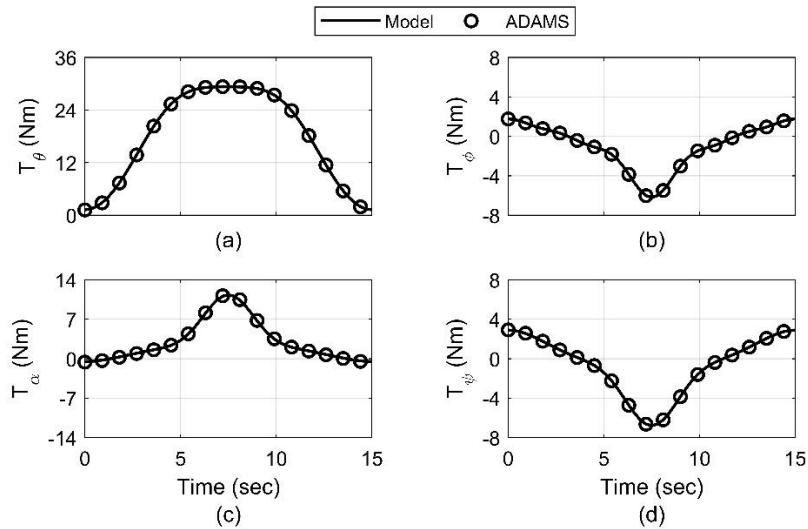

Figure 9. The mathematical model validation: (a) $T_\theta$ (b) $T_\varphi$ (c) $T_\alpha$ (d) $T_\psi$.

Strengthening trainings come after assistive trainings and can be performed in isometric or isotonic modes. During isometric mode, the robot stays at a fixed position while the patient applies force monitored by force sensors. During isotonic training, the robot applies a resistive force against the patient's motion. At the late stages of the rehabilitation process, patients may use other types of stationary-based robots to practice proprioceptive trainings such as balance exercises.

In order to demonstrate the feasibility of the proposed mechanism for the rehab trainings, the conventional rehab training, i.e. passive assistance, is chosen and examined. Therefore, a closed-loop control with position sensors (encoders) is sufficient to satisfy this requirement. Another reason for using closed-loop control is the uncertainty in the system parameters associated with different patients. Since subjects with different physical characteristics such as stature and weight are meant to use the device, a control strategy is required to ensure the fulfilment of the desired training. Therefore, a simple yet widely used controller in mechatronic systems, such as PI, seems appropriate for our preliminary investigation.

Before training, the ROM of each joint in which the joint can be moved passively is obtained experimentally. Then, the robot starts to cyclically move the joint based on the reference trajectory and within the calculated ROM. This ROM is increased gradually to recover the natural ROM.

For the control feedback loop to be simulated, a rearranged form of equations in (16) is required. The variation of $\psi$ during I/E as a function of $\phi$ obtained in the preliminary experiments (stated in the 'Results and Discussion' section) is used to omit $\psi$ in (16). Then, equations in (16) are solved for the second derivative of coordinates and S/P torque produced by foot ($\ddot{\theta}$, $\ddot{\varphi}$, $\ddot{\alpha}$ and $T_\psi$). The coordinates and their first derivatives can be obtained by integration of the second derivatives. Thus, we can obtain a model whose inputs are the three motor torques ($T_\theta$, $T_\varphi$ and $T_\alpha$), and outputs are the three independent coordinates ($\theta$, $\phi$ and $\alpha$). This model is used as the robot block in the simulations. Furthermore, the three motor torques are computed by the inverse dynamics of motors.

The reference trajectory used here is similar to the trajectory in the previous subsection; but this time, it moves between the two states at a constant speed (an isokinetic motion). At each state, the system remains still for a while to ensure muscle relaxation. This time, $\theta$ changes between 0° and 80°, and $\phi$ is limited to ±20°. However, the limits of $\alpha$ are rather unknown for I/E. These unknown quantities are determined by the preliminary experiments that declare the variations of $\alpha$ during I/E as a function of $\phi$ (this equation is also stated in the 'Results and Discussion' section).

To evaluate the performance of the controller in different situations, three subjects with different physical characteristics are considered in the simulations: subject 1 with 45 kg weight and 1.5 m height, subject 2 with 65 kg weight and 1.7 m height, and subject 3 with 85 kg weight and 1.9 m height.

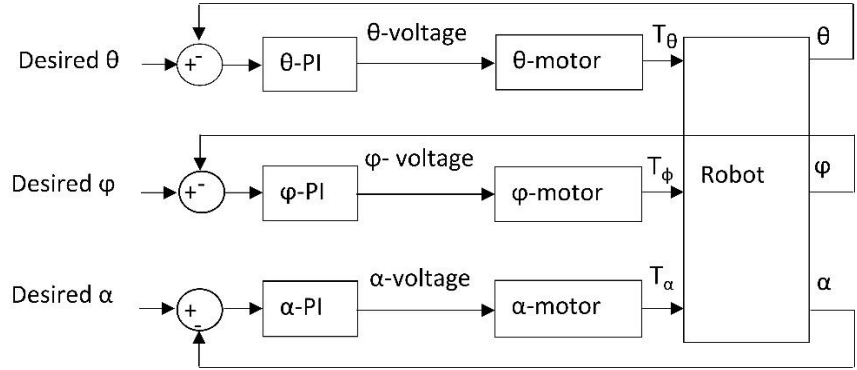

Figure 10. Block diagram of the control loop.

Figure 10 shows the control block diagram. Three PI controllers are used to control the motors that are tuned in MATLAB software. The gains obtained for θ-PI are $k_p$=120 and $k_i$=60, for ϕ-PI are $k_p$=4000 and $k_i$=60 and for α-PI are $k_p$=600 and $k_i$=80. The gains obtained here are later used in the experiments with some adjustments.

The results presented in Figure 11 demonstrate that the PI controller satisfies the demands of the rehab training adequately. The Root Mean Square (RMS) of the tracking error for θ-coordinate is less than 1.5°, while it is less than 0.6° for the other coordinates for all subjects. Thus, PI is a good candidate for practical tests.

## 5. Results and Discussion

### 5. 1 Test setup

Figure 12 shows a schematic layout of the test components. To implement the PI controllers, we used a microcontroller board (Arduino Duo, Arduino) in conjunction with motor drivers (L298, STMicroelectronics) and 200-pulse rotary optical encoders (HEDS-9700-E50, Avago Technologies). The controller calculates the required voltages for the motors based on the current position which are then delivered to the motor drivers as PWM signals. The drivers provide the required current for the motors.

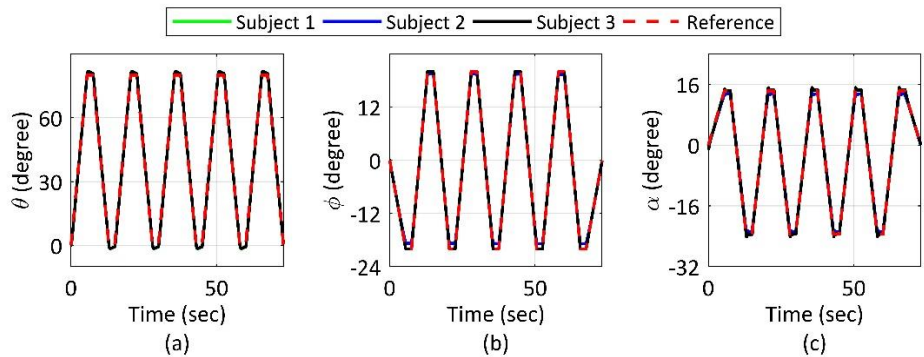

Figure 11. Control loop simulation results: (a) θ coordinate (b) ϕ coordinate (c) α coordinate.

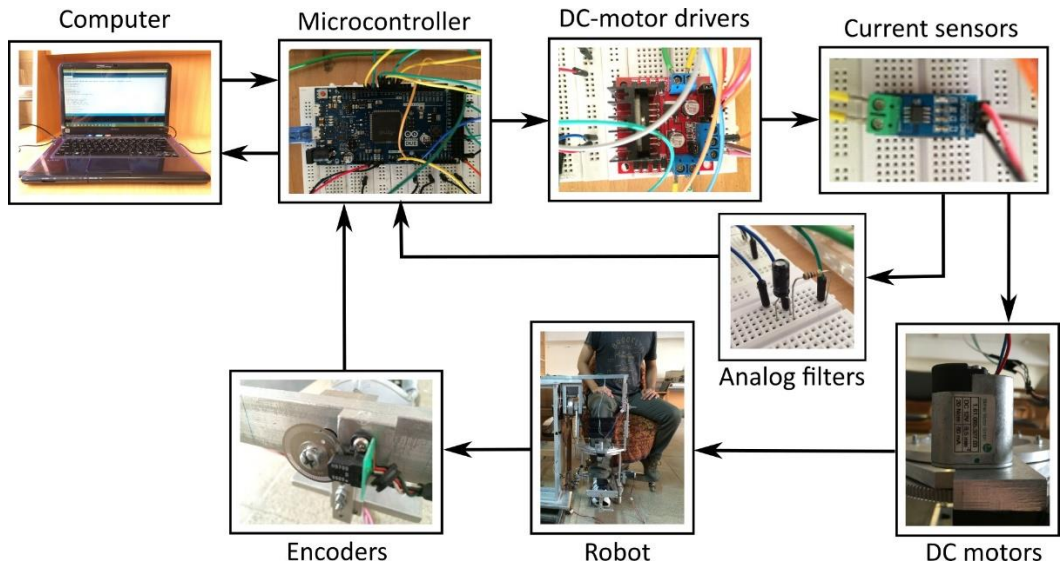

Figure 12. Test setup.

The θ-motor has been selected in previous work: a DC gear motor (ZKE2032-2K, Zhengk Electromotor) with 6 Nm rated torque and 55 rpm rated speed (torque constant≈0.5 Nm/A). Based on the simulation results, a DC geared motor (1.61.065.327, Bühler Motor) with 0.2 Nm rated torque and 9.5 rpm rated speed (torque constant≈3 Nm/A) is selected for actuation of A/A. Another DC geared motor (ZGB102FEE-30EE, Zhengk Electromotor) with 2.6 Nm rated torque and 9.1 rpm rated speed (torque constant≈6 Nm/A) is chosen to drive ankle P/D. To find the motor torques in practice, a current sensor (ASC712, Allegro MicroSystems) was employed in series with each motor. The outputs of current sensors were low-pass-filtered with 1.6 Hz cutoff-frequency and then multiplied by the motor torque constants in the microcontroller. All data are transferred to the computer for storage and display.

*5.2 Preliminary test*

A healthy subject volunteered to use the robot. His right leg was strapped to the robot. Informed consent was obtained based on the institutional ethical guidelines of the university. It is necessary to determine the ROMs and variations of $\alpha$ and $\psi$ as a function of $\phi$ during I/E through some preliminary tests. Therefore, the motors are detached and substituted with auxiliary shafts to enable the subject to move his foot freely. Then, the subject foot is moved between inversion and eversion; meanwhile, encoders record the rotations. Then, equations for $\alpha$ and $\psi$ (as a function of $\phi$) are obtained by linear estimation. These equations are then used to define the reference trajectory in both simulations and experiments and replacing $\psi$ with $\phi$ in (16).

In the preliminary test's result shown in Figure 13, it was found that (for the subject under training) the active ranges of $\phi$, $\alpha$ and $\psi$ change between 34°, -36° and 30° in inversion and -28°, 23° and -22° in eversion, respectively. For safety, $\phi$ is decided to be limited to ±20° in the reference trajectory. The angles $\psi$ and $\alpha$ (as a function of $\phi$ during I/E) are then approximated in Radian, as shown below:

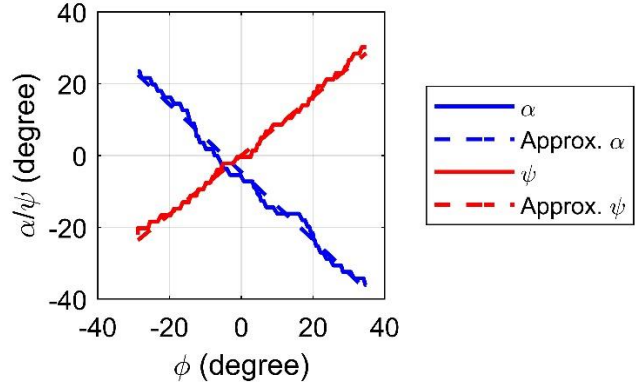

Figure 13. Preliminary test result.

$$\psi = 0.818\varphi$$
$$\alpha = -0.943\varphi - 0.08 \qquad (18)$$

*5.3 Test results*

The rehab training is now implemented using the proposed robot. During the rehab training, the subject is asked to make no effort. The reference trajectory is the same as the one used in the control loop simulation. However, the controller gains obtained in the simulations are used as an initial guess in the experiments. These gains are further modified in practice. We decrease the gain of the θ- and ϕ-PI by a factor of two, and the gain of α-PI by a factor of four. In other words, the gains are finally set to $k_p$=60 and $k_i$=30 for θ-PI, $k_p$=2000 and $k_i$=30 for ϕ-PI, and $k_p$=150 and $k_i$=20 for α-PI.

The rehab test results are presented in Figure 14. These results show how ankle-foot complex motion was realized in all anatomical planes during the rehab training (Figures 14c, e, g), while only two actuators were utilized for the actuation of I/E. Therefore, it is necessary for any rehab robot to provide foot with motions in the all three planes.

The simulated motor torques were presented in Figure 9. According to the simulations, the required θ-motor torque was about 29 Nm. This quantity was obtained higher in the experiments (around 31 Nm). This difference would be smaller if more accurate values for the system parameters were used. On the contrary, the differences were higher for ϕ- and α-motor torques. The reason could be that the subject's joints were rather compliant and must have had smaller stiffness than the one used in simulations. Furthermore, while the α-motor torque followed the trend in simulations, i.e. torque peaks were at dorsiflexion side, and there were negligible torque peaks in plantarflexion side, the ϕ-motor applied almost the same amount of torques in both A/A directions, which can be attributed to the friction in the custom-made thrust ball bearing of the robot.

As can be seen in Figures 11 and 14, there is good consistency between the experiment and simulation in terms of angular displacements. In fact, the PI controllers could achieve the objective of the rehab training very well. In passive assistance, it is critical for the robot to track the reference trajectory. The RMS values of errors for θ, ϕ and α coordinates in the test were 3.2°, 0.8° and 1.8°, respectively. These small errors between the reference trajectory and the experimental angular displacements assure that the requirements of passive assistance training have been met. Also, motion in ψ coordinate was observed during this experiment (Figure 14g) which is missing in the previous lower limb rehab robots [26], [28]–[32]. This result proves the coexistence of foot S/P and A/A in the passive motion of the foot. Therefore, these two DOFs should be considered simultaneously in rehab robot designs. Here, ROM of S/P was almost half of its value in the preliminary test (Figure 13); i.e., about 16° towards supination and 12° towards pronation directions. This result was expected because an external force moved patient foot during the preliminary test.

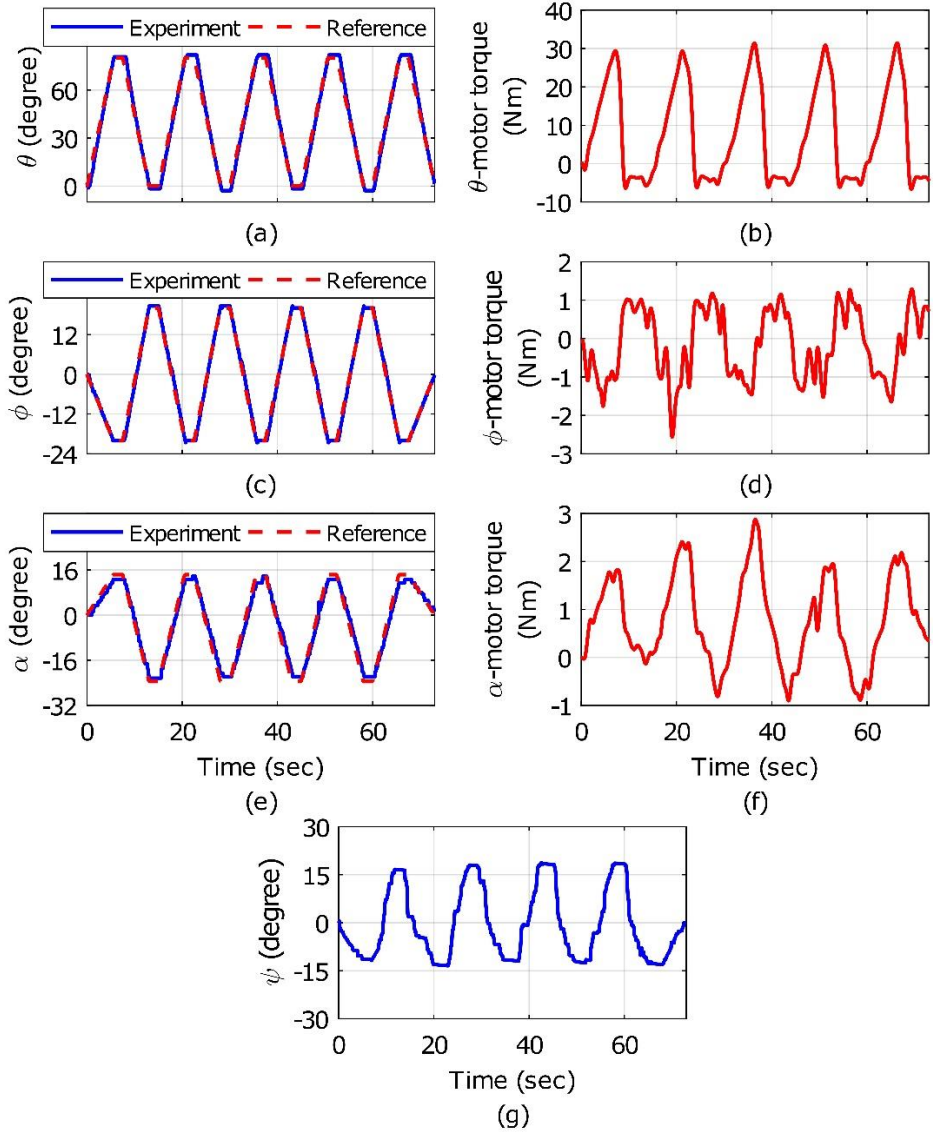

Figure 14. Test results: (a) θ coordinate (b) θ-motor torque (c) ϕ coordinate (d) ϕ-motor torque (e) α coordinate (f) α-motor torque (g) ψ coordinate.

## 6. Conclusion

A new stationary-based robot for simultaneous rehabilitation of the combination of knee, ankle and foot was presented in this work. Unlike previous robots, this robot comprises four DOFs: one for the knee F/E and three for the ankle-foot I/E, which generates I/E more naturally. Our results show the capability of the developed robot to conduct the common rehab trainings for increasing ROM and decreasing joint stiffness via passive assistance training.

Further work is to be conducted for the robot to be used in clinical setting. Handcuff-like rings should replace the A/A ball bearing for convenience. Besides, the robot should be made dynamically balanced around all rotation axes to facilitate the natural motion of lower limbs. The robot can also be equipped with force feedback loops to implement active assistance and strengthening isometric and isotonic trainings, and to account for erratic motions of patients.

**Declaration of interest**

The authors declare that there is no conflict of interest.

**Funding**

This research received no specific grant from any funding agency in the public, commercial, or not-for-profit sectors.

**References**

[1] R. Mehanna and J. Jankovic, "Movement disorders in cerebrovascular disease," *The Lancet Neurology*, vol. 12, no. 6, pp. 597–608, 2013.

[2] E. J. Benjamin *et al.*, "Heart Disease and Stroke Statistics—2017 Update: A Report From the American Heart Association," *Circulation*, 2017.

[3] W. Meng, Q. Liu, Z. Zhou, Q. Ai, B. Sheng, and S. S. Xie, "Recent development of mechanisms and control strategies for robot-assisted lower limb rehabilitation," *Mechatronics*, vol. 31, pp. 132–145, 2015.

[4] S. K. Banala, S. H. Kim, S. K. Agrawal, and J. P. Scholz, "Robot assisted gait training with active leg exoskeleton (ALEX)," in *2008 2nd IEEE RAS & EMBS International Conference on Biomedical Robotics and Biomechatronics*, 2008, pp. 653–658.

[5] S. Hussain, S. Q. Xie, P. K. Jamwal, and J. Parsons, "An intrinsically compliant robotic orthosis for treadmill training," *Medical engineering & physics*, vol. 34, no. 10, pp. 1448–1453, 2012.

[6] J. Yoon, B. Novandy, C.-H. Yoon, and K.-J. Park, "A 6-DOF gait rehabilitation robot with upper and lower limb connections that allows walking velocity updates on various terrains," *IEEE/ASME Transactions on Mechatronics*, vol. 15, no. 2, pp. 201–215, 2010.

[7] Y. Shao, Z. Xiang, H. Liu, and L. Li, "Conceptual design and dimensional synthesis of cam-linkage mechanisms for gait rehabilitation," *Mechanism and Machine Theory*, vol. 104, pp. 31–42, 2016.

[8] Y. Zhu *et al.*, "New wearable walking-type continuous passive motion device for

postsurgery walking rehabilitation," *Proceedings of the Institution of Mechanical Engineers, Part H: Journal of Engineering in Medicine*, vol. 227, no. 7, pp. 733–745, 2013.

[9] G. Chen, P. Qi, Z. Guo, and H. Yu, "Mechanical design and evaluation of a compact portable knee--ankle--foot robot for gait rehabilitation," *Mechanism and Machine Theory*, vol. 103, pp. 51–64, 2016.

[10] A. Roy *et al.*, "Robot-aided neurorehabilitation: a novel robot for ankle rehabilitation," *IEEE Transactions on Robotics*, vol. 25, no. 3, pp. 569–582, 2009.

[11] G. Belforte, G. Eula, S. Appendino, and S. Sirolli, "Pneumatic interactive gait rehabilitation orthosis: design and preliminary testing," *Proceedings of the Institution of Mechanical Engineers, Part H: Journal of Engineering in Medicine*, vol. 225, no. 2, pp. 158–169, 2011.

[12] F. Gao, Y. Ren, E. J. Roth, R. Harvey, and L.-Q. Zhang, "Effects of repeated ankle stretching on calf muscle–tendon and ankle biomechanical properties in stroke survivors," *Clinical biomechanics*, vol. 26, no. 5, pp. 516–522, 2011.

[13] S. W. O Driscoll and N. J. Giori, "Continuous passive motion (CPM): theory and principles of clinical application," *Journal of rehabilitation research and development*, vol. 37, no. 2, pp. 179–188, 2000.

[14] A. Alipour and M. J. Mahjoob, "A rehabilitation robot for continuous passive motion of foot inversion-eversion," in *2016 4th International Conference on Robotics and Mechatronics (ICROM)*, 2016, pp. 349–354.

[15] P. J. Mcnair, E. W. Dombroski, D. J. Hewson, S. N. Stanley, and others, "Stretching at the ankle joint: viscoelastic responses to holds and continuous passive motion," *Medicine & Science in Sports & Exercise*, vol. 33, no. 3, pp. 354–358, 2001.

[16] N. Hogan *et al.*, "Motions or muscles? Some behavioral factors underlying robotic assistance of motor recovery.," *Journal of rehabilitation research and development*, vol. 43 5, pp. 605–618, 2006.

[17] G. Aguirre-Ollinger, J. E. Colgate, M. A. Peshkin, and A. Goswami, "A one-degree-of-freedom assistive exoskeleton with inertia compensation: the effects on the agility of leg swing motion," *Proceedings of the Institution of Mechanical Engineers, Part H: Journal of Engineering in Medicine*, vol. 225, no. 3, pp. 228–245, 2011.

[18] G. Aguirre-Ollinger, "Exoskeleton control for lower-extremity assistance based on adaptive frequency oscillators: Adaptation of muscle activation and movement frequency," *Proceedings of the Institution of Mechanical Engineers, Part H: Journal of Engineering in Medicine*, vol. 229, no. 1, pp. 52–68, 2015.

[19] N. Naghavi and M. J. Mahjoob, "Design and control of an active 1-DoF mechanism for knee rehabilitation," *Disability and Rehabilitation: Assistive Technology*, no. 0, pp. 1–7, 2015.

[20] M. Girone, G. Burdea, M. Bouzit, V. Popescu, and J. E. Deutsch, "A Stewart platform-based system for ankle telerehabilitation," *Autonomous Robots*, vol. 10, no. 2, pp. 203–212, 2001.

[21] Jungwon Yoon and Jeha Ryu, "A Novel Reconfigurable Ankle/Foot Rehabilitation Robot," in *Proceedings of the 2005 IEEE International Conference on Robotics and Automation*,

2005, pp. 2290–2295.

[22] Y. H. Tsoi and S. Q. Xie, "Design and Control of a Parallel Robot for Ankle Rehabiltation," in *2008 15th International Conference on Mechatronics and Machine Vision in Practice*, 2008, pp. 515–520.

[23] J. A. Saglia, N. G. Tsagarakis, J. S. Dai, and D. G. Caldwell, "A high performance 2-dof over-actuated parallel mechanism for ankle rehabilitation," in *IEEE International Conference on Robotics and Automation (ICRA)*, 2009, pp. 2180–2186.

[24] M. Zhang, J. Cao, G. Zhu, Q. Miao, X. Zeng, and S. Q. Xie, "Reconfigurable workspace and torque capacity of a compliant ankle rehabilitation robot (CARR)," *Robotics and Autonomous Systems*, vol. 98, pp. 213–221, 2017.

[25] D. E. Voss, M. K. Ionta, B. J. Myers, and M. Knott, *Proprioceptive neuromuscular facilitation: patterns and techniques*. Harper & Row Philadelphia, PA, 1985.

[26] M. Bouri, B. Le Gall, and R. Clavel, "A new concept of parallel robot for rehabilitation and fitness: The Lambda," in *2009 IEEE International Conference on Robotics and Biomimetics (ROBIO)*, 2009, pp. 2503–2508.

[27] E. Akdoğan and M. A. Adli, "The design and control of a therapeutic exercise robot for lower limb rehabilitation: Physiotherabot," *Mechatronics*, vol. 21, no. 3, pp. 509–522, 2011.

[28] V. Monaco, G. Galardi, M. Coscia, D. Martelli, and S. Micera, "Design and Evaluation of NEUROBike: A Neurorehabilitative Platform for Bedridden Post-Stroke Patients," *IEEE Transactions on Neural Systems and Rehabilitation Engineering*, vol. 20, no. 6, pp. 845–852, Nov. 2012.

[29] W. Wang, Z.-G. Hou, L. Tong, F. Zhang, Y. Chen, and M. Tan, "A novel leg orthosis for lower limb rehabilitation robots of the sitting/lying type," *Mechanism and Machine Theory*, vol. 74, pp. 337–353, 2014.

[30] J. Wu, J. Gao, R. Song, R. Li, Y. Li, and L. Jiang, "The design and control of a 3DOF lower limb rehabilitation robot," *Mechatronics*, 2016.

[31] S. Mohan, J. K. Mohanta, S. Kurtenbach, J. Paris, B. Corves, and M. Huesing, "Design, development and control of a 2PRP-2PPR planar parallel manipulator for lower limb rehabilitation therapies," *Mechanism and Machine Theory*, vol. 112, pp. 272–294, 2017.

[32] J. K. Mohanta, S. Mohan, P. Deepasundar, and R. Kiruba-Shankar, "Development and control of a new sitting-type lower limb rehabilitation robot," *Computers & Electrical Engineering*, vol. 67, pp. 330–347, 2018.

[33] Z. Zhou, Y. Zhou, N. Wang, F. Gao, K. Wei, and Q. Wang, "A proprioceptive neuromuscular facilitation integrated robotic ankle–foot system for post stroke rehabilitation," *Robotics and Autonomous Systems*, 2014.

[34] S. Siegler, J. Chen, and C. D. Schneck, "The three-dimensional kinematics and flexibility characteristics of the human ankle and subtalar joints—Part I: Kinematics," *Journal of biomechanical engineering*, vol. 110, no. 4, pp. 364–373, 1988.

[35] A. J. van den Bogert, G. D. Smith, and B. M. Nigg, "In vivo determination of the anatomical axes of the ankle joint complex: an optimization approach," *Journal of biomechanics*, vol.

27, no. 12, pp. 1477–1488, 1994.

[36] N. Sancisi, B. Baldisserri, V. Parenti-Castelli, C. Belvedere, and A. Leardini, "One-degree-of-freedom spherical model for the passive motion of the human ankle joint," *Medical & biological engineering & computing*, vol. 52, no. 4, pp. 363–373, 2014.

[37] I. A. Kapandji, *Physiology of the Joints: Lower Limb: Volume 2*, 6th ed. Churchill Livingstone, 2010.

[38] S. S. Adler, D. Beckers, and M. Buck, *PNF in practice: an illustrated guide*. Springer, 2007.

[39] S. Zographos, B. Chaminade, M. C. Hobatho, and G. Utheza, "Experimental study of the subtalar joint axis preliminary investigation," *Surgical and Radiologic Anatomy*, vol. 22, no. 5–6, pp. 271–276, 2001.

[40] Y. Ding, M. Sivak, B. Weinberg, C. Mavroidis, and M. K. Holden, "NUVABAT: northeastern university virtual ankle and balance trainer," in *IEEE Haptics Symposium*, 2010, pp. 509–514.

[41] T.-W. Lu and J. J. O'connor, "Bone position estimation from skin marker co-ordinates using global optimisation with joint constraints," *Journal of biomechanics*, vol. 32, no. 2, pp. 129–134, 1999.

[42] R. Storn and K. Price, "Differential evolution--a simple and efficient heuristic for global optimization over continuous spaces," *Journal of global optimization*, vol. 11, no. 4, pp. 341–359, 1997.

[43] S. H. Kafash and A. Nahvi, "Optimal synthesis of four-bar path generator linkages using Circular Proximity Function," *Mechanism and Machine Theory*, vol. 115, pp. 18–34, 2017.

[44] M. L. Latash and V. M. Zatsiorsky, "Joint stiffness: Myth or reality?," *Human movement science*, vol. 12, no. 6, pp. 653–692, 1993.

[45] N. Sancisi, D. Zannoli, V. Parenti-Castelli, C. Belvedere, and A. Leardini, "A one-degree-of-freedom spherical mechanism for human knee joint modelling," *Proceedings of the Institution of Mechanical Engineers, Part H: Journal of Engineering in Medicine*, vol. 225, no. 8, pp. 725–735, 2011.

[46] C. A. Oatis, "The use of a mechanical model to describe the stiffness and damping characteristics of the knee joint in healthy adults," *Physical Therapy*, vol. 73, no. 11, pp. 740–749, 1993.

[47] J. Chen, S. Siegler, and C. D. Schneck, "The three-dimensional kinematics and flexibility characteristics of the human ankle and subtalar joint—part II: flexibility characteristics," *Journal of biomechanical engineering*, vol. 110, no. 4, pp. 374–385, 1988.

[48] D. A. Winter, *Biomechanics and motor control of human movement*. John Wiley & Sons, 2009.

[49] J. A. Saglia, N. G. Tsagarakis, J. S. Dai, and D. G. Caldwell, "Control strategies for patient-assisted training using the ankle rehabilitation robot (ARBOT)," *IEEE/ASME Transactions on Mechatronics*, vol. 18, no. 6, pp. 1799–1808, 2012.

[50] J. Dul and G. E. Johnson, "A kinematic model of the human ankle," *Journal of Biomedical Engineering*, vol. 7, no. 2, pp. 137–143, 1985.